\documentclass{article}
\usepackage{amsmath,epsfig}
\usepackage[preprint]{spconfa4}
\let\OLDthebibliography\thebibliography
\copyrightnotice{Copyright notice – please select from the list provided at Camera-Ready Instructions}
\renewcommand\thebibliography[1]{
  \OLDthebibliography{#1}
  \setlength{\parskip}{0pt}
  \setlength{\itemsep}{0pt plus 0.3ex}
}

\usepackage[hyphens]{url}
\usepackage{algorithm}
\usepackage{algorithmic}
\usepackage{newfloat}
\usepackage{listings}
\usepackage{bm}
\usepackage{amssymb}
\usepackage{booktabs}
\usepackage{multirow}
\usepackage[table,xcdraw]{xcolor}
\usepackage{amsmath}
\usepackage{xcolor}
\usepackage{subcaption} 
\usepackage{pifont}
\newcommand{\cmark}{\ding{51}}%
\newcommand{\xmark}{\ding{55}}%

\def\bx{\mathbf{x}}
\def\bw{\mathbf{w}}
\def\bz{\mathbf{z}}
\def\bv{\mathbf{v}}
\def\DD{\mathcal{D}}

\begin{document}

\title{FEDIC: Federated Learning on Non-IID and Long-Tailed Data via Calibrated Distillation}
%
\name{Xinyi Shang$^{\rm 1}$, Yang Lu$^{\rm 1,\dagger}$, Yiu-ming Cheung$^{\rm 2}$, Hanzi Wang$^{\rm 1}$ \thanks{$^{\dagger}$Corresponding author: Yang Lu, luyang@xmu.edu.cn}\thanks{This work was supported in part by the National Natural Science Foundation of China under Grants 62002302, 61872307, 61672444 and U21A20514; in part by the Open Research Projects of Zhejiang Lab (NO. 2021KB0AB03); in part by the National Natural Science Foundation of China/RGC Joint Research Scheme under Grant N\_HKBU214/21; in part by the Natural Science Foundation of Fujian Province under Grant 2020J01005.}}
\address{$^{\rm 1}$Xiamen University, $^{\rm 2}$Hong Kong Baptist University\\$^{\rm 1}$shangxinyi@stu.xmu.edu.cn, $^{\rm 1}$\{luyang, hanzi.wang\}@xmu.edu.cn, $^{\rm 2}$ymc@comp.hkbu.edu.hk}

\maketitle
\begin{abstract}
Federated learning provides a privacy guarantee for generating good deep learning models on distributed clients with different kinds of data. Nevertheless, dealing with non-IID data is one of the most challenging problems for federated learning. Researchers have proposed a variety of methods to eliminate the negative influence of non-IIDness. However, they only focus on the non-IID data provided that the universal class distribution is balanced. In many real-world applications, the universal class distribution is long-tailed, which causes the model seriously biased. Therefore, this paper studies the joint problem of non-IID and long-tailed data in federated learning and proposes a corresponding solution called Federated Ensemble Distillation with Imbalance Calibration (FEDIC). To deal with non-IID data, FEDIC uses model ensemble to take advantage of the diversity of models trained on non-IID data. Then, a new distillation method with logit adjustment and calibration gating network is proposed to solve the long-tail problem effectively. We evaluate FEDIC on CIFAR-10-LT, CIFAR-100-LT, and ImageNet-LT with a highly non-IID experimental setting, in comparison with the state-of-the-art methods of federated learning and long-tail learning. 
Our code is available at \color{magenta}\url{https://github.com/shangxinyi/FEDIC}\color{black}. 
\end{abstract}
\vspace{-5px}
\section{Introduction}
\vspace{-5px}
\begin{figure}[!t]
	\centering
	\includegraphics[width=0.8\linewidth]{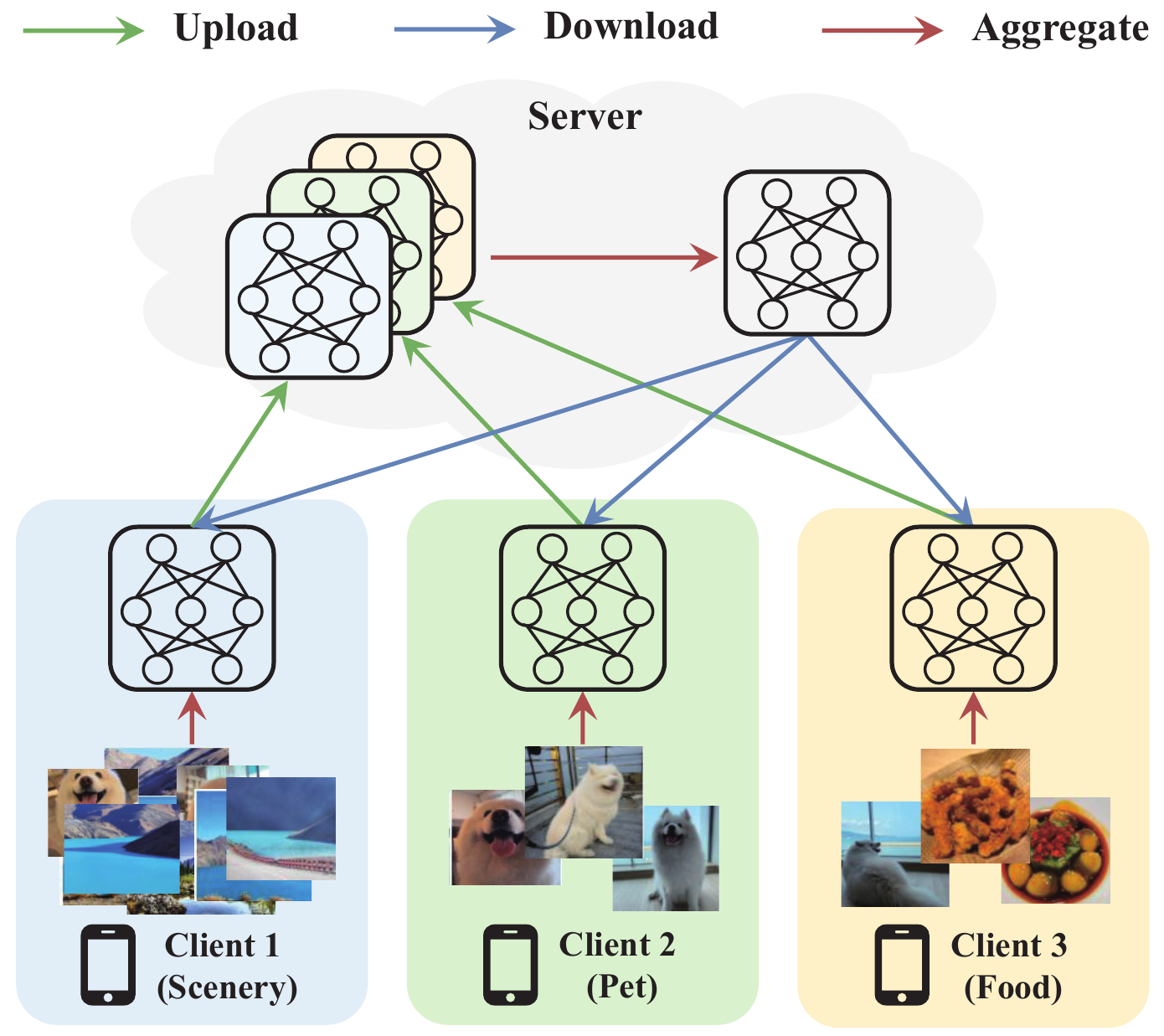}
	\caption{An example of federated learning application for the task of gallery tagging on mobile phones. The majority class in each client is shown in parentheses.
	}
	\label{example}
	\vspace{-12px}
\end{figure}
In recent years, an increasing number of deep learning techniques have been deployed in mobile devices to handle data from different sources, e.g., cameras, microphones, GPS, and other sensors. These data play a key role in generating strong predictive models to provide better services to the users. However, transmitting user data to the server would bring high privacy risks for both the service providers and mobile users \cite{9102960}. Recently, federated learning has received increasing attention 
due to its capacity for distributed machine learning with privacy protection \cite{li2020federated}. Data privacy is guaranteed by storing data and training model locally on each client. 
The global model on the server is produced by aggregating local models transmitted from clients without the requirement of any data from them \cite{mcmahan2017communication}.
However, data heterogeneity is still a major challenge in federated learning. 
Since 
the data in each client may be drawn from different distributions without meeting the requirement of IID \cite{zhao2018federated}, 
training on this kind of data results in poor generalization ability of the global model.

In the literature, a number of methods have been proposed to deal with non-IID data in federated learning. They can be roughly categorized into client-side methods and server-side methods, respectively. The former aims to improve the local training process. Most of them regularize the local training process such that the diversity of client models can be limited \cite{MLSYS2020_38af8613}. The latter adopts specific model aggregation mechanisms to alleviate the negative influence of data heterogeneity \cite{9428075, 9428189}. Some recent works have adopted knowledge distillation on the server \cite{ NEURIPS2020_18df51b9,chen2021fedbe}. The knowledge is transfered from an ensemble model, which is built by local models, to the global model. There are also some other methods that focus on optimization strategy \cite{NEURIPS2020_564127c0, hsu2019measuring} on the server.

Although the abovementioned methods solve the data non-IIDness problem to some extent, they generally assume that the universal class distribution is balanced, 
which may not be true from a practical perspective. 
As shown in Fig. \ref{example}, if we consider the overall clients, a few classes like scenery have a large number of samples, while many classes like pet or food only take a small portion. Building a classification model on this kind of distribution is termed long-tail learning \cite{zhang2021deep}, which has been extensively studied in recent years. Some methods origin from the traditional imbalance learning \cite{he2009learning}, where the re-sampling \cite{chou2020remix} or re-weighing \cite{ NEURIPS2019_e58cc5ca} techniques are adopted to alleviate the imbalance influence. The other methods \cite{Kang2020Decoupling} take advantage of the uniqueness of the deep learning model and focus on the representation learning. 

The existing solutions for non-IID data in federated learning generally perform poorly on the tail classes due to the lack of consideration of the universal long-tail distribution. The global class distribution is long-tailed such that each client only holds a few tail classes, which makes local models perform poorly on the tail classes. Therefore, the global model aggregated by biased local models is also biased.
There are also some methods specifically designed for federated learning on imbalanced data. One strategy is to adopt client selection to match complementary clients \cite{duan2020self}. However, some clients may lose the chance to participate in model aggregation on the server if they cannot be matched with other clients. Recently, ratio loss \cite{wang2021addressing} has been proposed to estimate the global imbalance status to help improve local optimization. However, the performance of ratio loss is dropped as the degree of data non-IIDness increases.

In this paper, we study the problem of federated learning on non-IID and long-tailed data, and correspondingly propose an effective server-side method called Federated Ensemble Distillation with Imbalance Calibration (FEDIC) without prior knowledge of global class distribution. In FEDIC, the knowledge distillation technique is adopted on the server to transfer the knowledge from the ensemble model to the global model. 
However, in the long-tailed setting, the ensemble model may still be biased towards the head classes. Subsequently, the transferred knowledge may not be helpful.
We therefore propose a novel ensemble calibration method to eliminate the bias of the ensemble model before conducting knowledge distillation. 
Specifically, we first propose a new logit adjustment to reconstruct the ensemble model from the perspectives of clients and classes, respectively. Then, a calibration gating network is proposed to fuse the adjusted logits based on ensemble representations effectively. The final ensemble model generalizes well on both head and tail classes after calibration.
The contributions of this paper can be summarized as follows:
\begin{itemize}
	\vspace{-5px}
	\item This paper is a pioneering work in federated learning to study the joint problems of the data non-IIDness and long-tail learning.
	\vspace{-5px}
	\item We propose a new ensemble calibration method by logit adjustment and calibration gating network techniques to effectively make the output of the ensemble model unbiased.
	\vspace{-5px}
	\item We propose an effective server-side federated learning method FEDIC, which utilizes the knowledge distillation technique to enhance the robustness of the global model on both non-IID and long-tailed data.
\end{itemize}
\setlength{\belowcaptionskip}{-15px}
\begin{figure*}[!t]
	\centering
	\includegraphics[width=\linewidth]{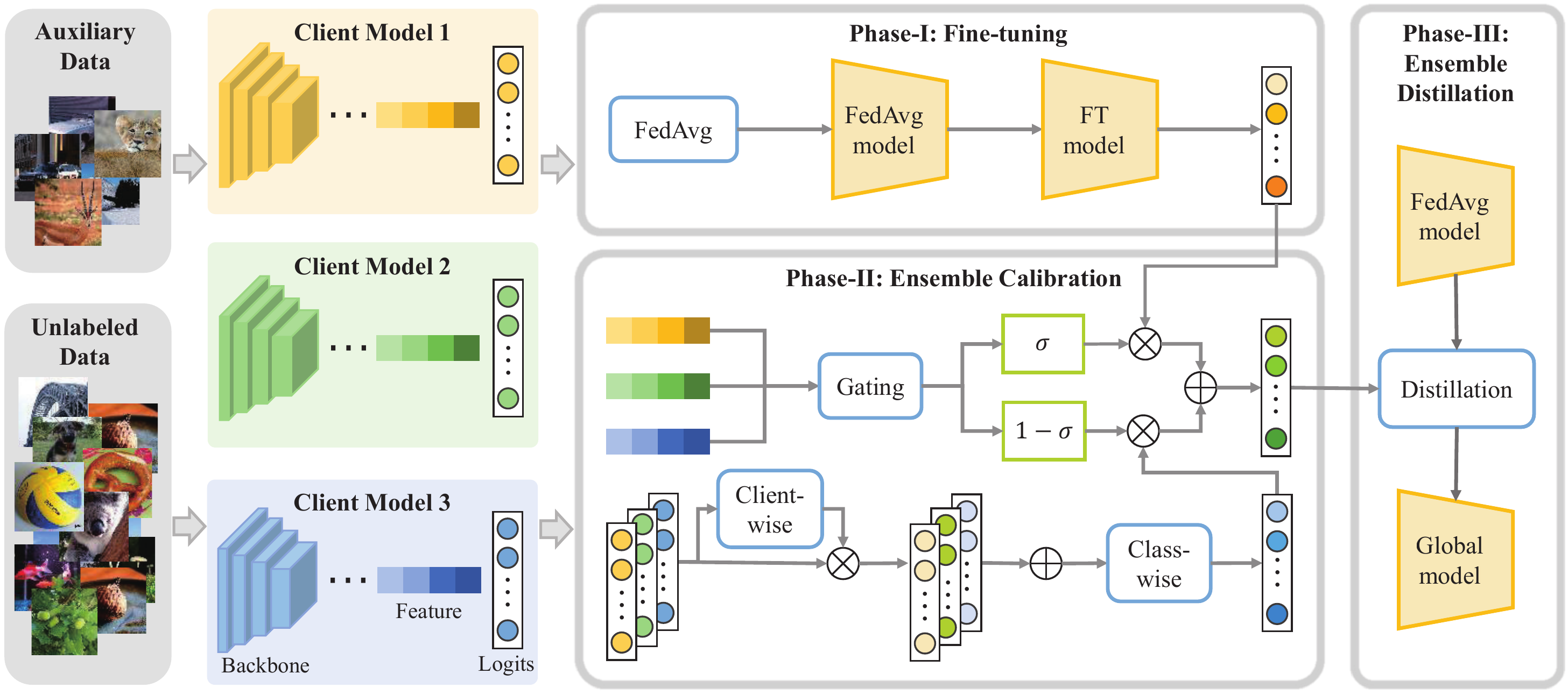}
	\caption{The framework of FEDIC on the server.
	}
	\label{model_figure}
\end{figure*}
\section{Proposed method}
\vspace{-5px}
In this section, we first describe the problem setting with some basic notations and then introduce FEDIC for federated learning on non-IID and long-tailed data.
\vspace{-5px}
\subsection{Problem Setting}
\vspace{-5px}
In this paper, the learning scenario is based on a typical federated learning system with $K$ clients holding potentially non-IID local datasets $\DD^1, \DD^2, ..., \DD^K$, respectively. The goal is to obtain a global model on the server over the union of all these datasets $\DD\triangleq \bigcup_{k\leq K} \DD^k$ without access to any data $\DD^k$ on the $k$-th client. The setting difference in this paper is that $\DD$ is drawn from a long-tailed distribution $(\mathcal{X},\mathcal{Y})$, $\mathcal{Y}\in\{1,...,C\}$, which is unknown in advance. The model in federated learning is typically a neural network $\phi_{\bw}$ with parameters $\bw$. $\phi_{\bw}$ has two components: 1) a feature extractor $f_{\bw}$, mapping each sample $\bx$ to a $d$-dim representation vector; 2) a classifier $h_{\bw}$, typically being a fully-connected layer which outputs logits to denote class confidence scores. The parameters of client $k$'s local model are denoted as $\bw_k$.

\vspace{-5px}
\subsection{Proposed Method}
\vspace{-5px}
FEDIC is a server-side method based on FedAvg \cite{mcmahan2017communication} without intervening in the local training process on each client. It is based on an intuitive idea: The ensemble of the local client models has better generalization ability than the global model produced by parameter averaging \cite{NEURIPS2020_18df51b9}. Since the local models are trained on non-IID data, their prediction results are highly diverse, which is one of the most important factors that make the ensemble model work better than a single model \cite{dietterich2000ensemble}. However, due to model heterogeneity, the ensemble model cannot be transmitted to the clients for further updating. Therefore, it is straightforward to leverage knowledge distillation \cite{hinton2015distilling} to transfer the generalization ability from the ensemble model to the global model. Then, the distilled global model is transmitted to each client for further updating. 

Specifically, on the server, we can construct the ensemble model as the teacher model:
\setlength{\abovedisplayskip}{3pt}
\setlength{\belowdisplayskip}{3pt}
\begin{align}\label{teacher}
\phi^t(\bx)=\sum_{k=1}^Ke_k\phi_{\bw_k}(\bx),
\end{align}
where $e_k$ is the ensemble weight for client $k$'s local model. Then, we can obtain the global model $\bw$ as the corresponding student model \footnote{For better notation representation, we ignore the superscript $(t)$ to denote the model in the $t$-th round, which is usually adopted in federated learning literature.} \cite{mcmahan2017communication}:
\begin{align}
\bw &= \sum_{k=1}^K\frac{|\DD^k|}{|\DD|}\bw_k,\\
\phi^s(\bx) &= \phi_{\bw}(\bx).\label{student}
\end{align}
However, when the universal class distribution is long-tailed, the generalization ability of the ensemble model on the tail classes is also poor. 
As a result,it cannot provide a helpful guide to the student model on the tail classes. 
Therefore, we propose to utilize a small auxiliary dataset $\DD_{aux}$ on the server, which is labeled and balanced in order to calibrate the ensemble model against the long-tail distribution. The main reason of utilizing the auxiliary data is that the global imbalance degree is unknown for both the server and clients, which makes most of the methods for long-tailed learning are infeasible. 
It is worth noting that all auxiliary data is collected independently on the server and there is no data transmission or data sharing in our problem setting.
The model architecture of FEDIC on the server is shown in Fig. \ref{model_figure}. In the following, we will describe two core components of FEDIC in detail.


\begin{table*}[!t]\footnotesize
	\centering
	\caption{Top-1 test accuracy ($\%$) for FEDIC and compared FL methods on CIFAR-10/100-LT with different IFs.}
	\begin{tabular}{@{}llcccccc@{}}
		\toprule
		&                    & \multicolumn{3}{c}{CIFAR-10-LT}                                & \multicolumn{3}{c}{CIFAR-100-LT}                               \\ \cmidrule(l){3-8} 
		\multirow{-2}{*}{\textbf{Family}}        & \multirow{-2}{*}{\textbf{Method}}   & \multicolumn{1}{c}{IF=100}    & \multicolumn{1}{c}{IF=50}    & \multicolumn{1}{c}{IF=10}    & \multicolumn{1}{c}{IF=100}    & \multicolumn{1}{c}{IF=50}    & \multicolumn{1}{c}{IF=10}    \\ \midrule
		& FedAvg                 & 52.12             & 52.43             & 59.97             & 25.81             & 28.19             & 38.22             \\
		& FedAvgM                & 53.64             & 54.42             & 59.52             & 25.11             & 28.82             & 38.77             \\
		&FedProx  & 52.75             & 55.07             & 60.44             & 25.43             & 27.77             & 38.45 \\
		\multirow{-4}{*}{FL methods}          & FedNova                & 52.93             & 56.53             & 61.58             & 26.81             & 28.91             & 39.62             \\ \midrule
		& FedDF                 & 50.33             & 52.58             & 58.84             & 25.60             &              28.79             &           38.60               \\
		\multirow{-2}{*}{Distillation-based FL methods} & FedBE                 &          44.05             & 
		50.66             &            53.53             &            22.46             &       
		23.77             &
		33.53           \\ \midrule
		& Fed-Focal Loss           & 49.66             & 52.02             & 59.68             & 24.66             & 26.04             & 35.54             \\
		& Ratio Loss         &           
		54.15             &  
		57.77             &            
		60.58             &
		26.72             &  
		28.83             & 
		38.79        \\
		& FedAvg+cRT               & 51.74             & 55.87             & 61.11             & 30.73             &              31.47             &  
		39.75      \\
		& FedAvg+$\tau$-norm           & 
		44.38   & 
		45.59   & 
		48.29   & 
		19.59   & 
		22.07   & 
		30.48   \\
		\multirow{-5}{*}{Imbalance-oriented FL methods} & FedAvg+LWS               & 44.48             & 46.20             & 55.17             & 20.70             &              23.24             &  
		32.31              \\ \midrule
		\multirow{-1}{*}{\cellcolor[HTML]{EFEFEF}Proposed method}
		& \cellcolor[HTML]{EFEFEF}FEDIC & \cellcolor[HTML]{EFEFEF}\textbf{63.11} & \cellcolor[HTML]{EFEFEF}\textbf{63.82} & \cellcolor[HTML]{EFEFEF}\textbf{65.50} & \cellcolor[HTML]{EFEFEF}\textbf{33.67} & \cellcolor[HTML]{EFEFEF}\textbf{34.74} & \cellcolor[HTML]{EFEFEF}\textbf{41.93} 
		\\ \bottomrule
	\end{tabular}
	\label{t1}
	\vspace{-12px}
\end{table*}
\noindent\textbf{Ensemble Calibration.} Because of training on local data with different distributions, each local model may perform differently on the tail classes. It is reasonable to assign a higher ensemble weight to the local model that performs well on the tail classes to improve the generalization ability of the ensemble model. However, the server has no prior knowledge of which classes are the tail classes and which local model performs well on them. 
Therefore, instead of giving each client a static weight (e.g., $1/K$ for the common average ensemble) in the ensemble, we propose the client-wise logit adjustment that searches proper ensemble weights $e_k, k=1,...,K$ by learnable parameters. Given a sample $\bx\in\DD_{aux}$ on the server, we first calculate the logits of local models $\phi_{\bw_k}(\bx)$. The ensemble weights $e_k$ are calculated by a non-linear transform:
\begin{align}\label{ensemble_weight_calculation}
e_k&=\mathrm{sigmoid}\big(\mathbf{a}_e^T\phi_{\bw_k}(\bx) + b_e\big),
\end{align}
where $\mathbf{a}_e\in\mathbb{R}^C$ and $b_e$ is a learnable parameter. $e_k$ is then normalized to make its sum equal to 1. 
Subsequently, the weighted logits of the local model can be computed, as shown in Eq. (\ref{teacher}). 
However, if none of the clients handles the tail classes well, the weighted ensemble is still biased towards the head classes. Subsequently, we propose class-wise logit adjustment to further enhance the logit of the tail classes by learnable parameters $\mathbf{a}_z,\mathbf{b}_z\in\mathbb{R}^C$. They linearly transform the original weighted ensemble logits $\phi^t(\bx)$ to calibrated logits $\bz^{cl}$ on each class:
\begin{align}\label{eq.2}
\bz^{cl}=\mathbf{a}_z\odot \phi^t(\bx) + \mathbf{b}_z,
\end{align}
where $\odot$ denotes the Hadamard product. Thus, $\bz^{cl}$ is the calibrated logits after client-wise and class-wise logit adjustment. 

However, the effectiveness of the premise of logit adjustment is that the features are well extracted. Simply manipulating the logits may not be sufficient if the feature extractors of local models are severely affected.
Therefore, we propose to update the feature extractor as well to complement logit adjustment. Specifically, we can obtain a model $\widehat\bw$ by fine-tuning the global model on $\DD_{aux}$. Since $\DD_{aux}$ is balanced, $\widehat\bw$ is adjusted to obtain an unbiased feature extractor. Then, we can obtain the fine-tuned logits $\bz^{ft}=\phi_{\widehat\bw}(\bx)$ for the input $\bx$.
The logits $\bz^{cl}$ and $\bz^{ft}$ are both adjusted to deal with the long-tail distribution but they are from different perspectives. That is, $\bz^{cl}$ is produced on the merit of the model ensemble but with fixed feature extractors, while $\bz^{ft}$ is based on the single global model, but its feature extractor is fine-tuned on $\DD_{aux}$. Inspired by \cite{zhang2021distribution}, we propose a calibration gating network to control the trade-off between $\bz^{ft}$ and $\bz^{cl}$, in order to effectively integrate the calibrated and fine-tuned logits and make them complement each other. The network takes the feature ensemble as the input through a non-linear layer to output the weight between $\bz^{ft}$ and $\bz^{cl}$, such that each sample obtains a different weight according to its own feature. The calibration gating network is formulated as:
\begin{align}
\sigma=\mathrm{sigmoid}(\mathbf{u}^T\bv),
\end{align}
where $\bv=\frac{1}{|S|}{\sum_{k\in S}f_{\bw_k}(\bx)}$ is the feature ensemble, and $|S|$ is the number of selected clients in each round. $\mathbf{u}\in\mathbb{R}^{d}$ is a learnable parameter.
Thus, the final calibrated logits $\bz'$ through the calibration gating network is formulated as:
\begin{align} \label{weight}
\bz'=\sigma \bz^{cl} + \big(1-\sigma\big)\bz^{ft}.
\end{align}
The weight $\sigma\in(0,1)$ acts as a feature-dependent gate to control the trade-off between $\bz^{ft}$ and $\bz^{cl}$. All learnable parameters in the whole process of ensemble calibration are updated by cross-entropy loss on $(\bx,\mathbf{y})\thicksim\DD_{aux}$:
\begin{align}\label{calibration_update}
L=-\sum_{j=1}^C y_j\log\frac{\exp(z_j')}{\sum_{i=1}^C\exp(z'_k)}.
\end{align}

\noindent\textbf{Ensemble Distillation.} To better distill unbiased knowledge from the teacher model (i.e., the calibrated ensemble model) to the student model (i.e., the global model), we follow the work of knowledge distillation with two loss components \cite{hinton2015distilling}: (1) $L_{CE}$ is the cross-entropy loss between logits of the student model and the ground truth; (2) $L_{KL}$ is the Kullback-Leibler (KL) divergence of the logits between the teacher model and the student model. We use $\DD_{aux}$ to calculate $L_{CE}$, and use another unlabeled dataset $\DD_{ulb}$ for $L_{KL}$ to boost the performance of distillation knowledge further. Thus, the loss is constituted by a trade-off hyperparameter $\lambda\in[0,1]$:
\begin{align}\label{distill}
L_{FEDIC}&=(1-\lambda)L_{CE}+\lambda L_{KL}.
\end{align}
We set $\lambda=0.5$ in all experiments. 
\vspace{-5px}
\section{Experiments}
\vspace{-5px}
\subsection{Experiment Setup} \label{setting}
\vspace{-5px}
\begin{figure*}[!t]
	\centering
	\begin{minipage}[t]{0.31\linewidth}
		\centering
		\includegraphics[width=\linewidth]{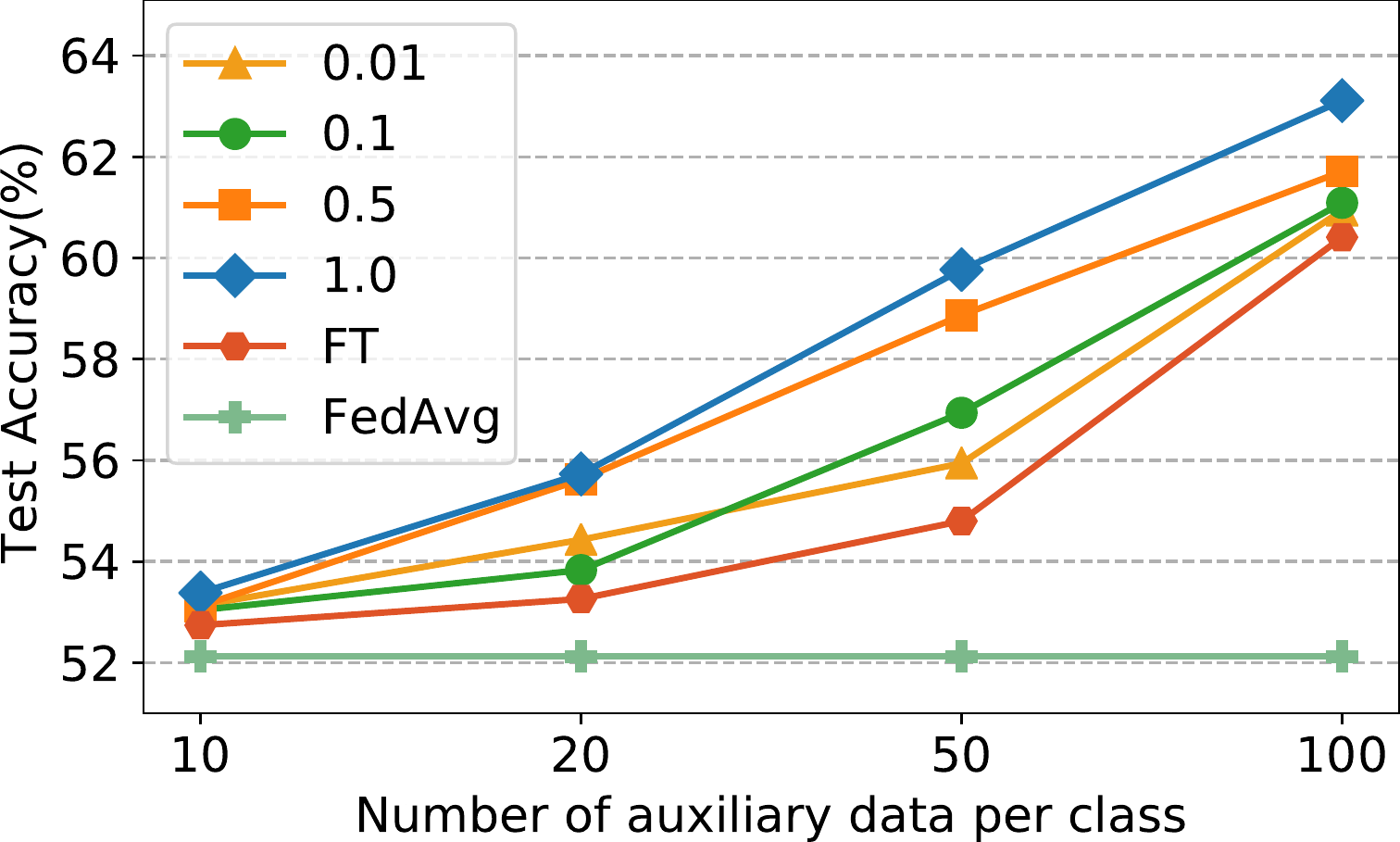}
		\caption{The performance of FEDIC with different sizes of auxiliary data on CIFAR-10-LT with IF=100.}
		\label{fig_unlabeled}
	\end{minipage}
	\hspace{0.08in} 
	\begin{minipage}[t]{0.31\linewidth}
		\centering
		\includegraphics[width=\linewidth]{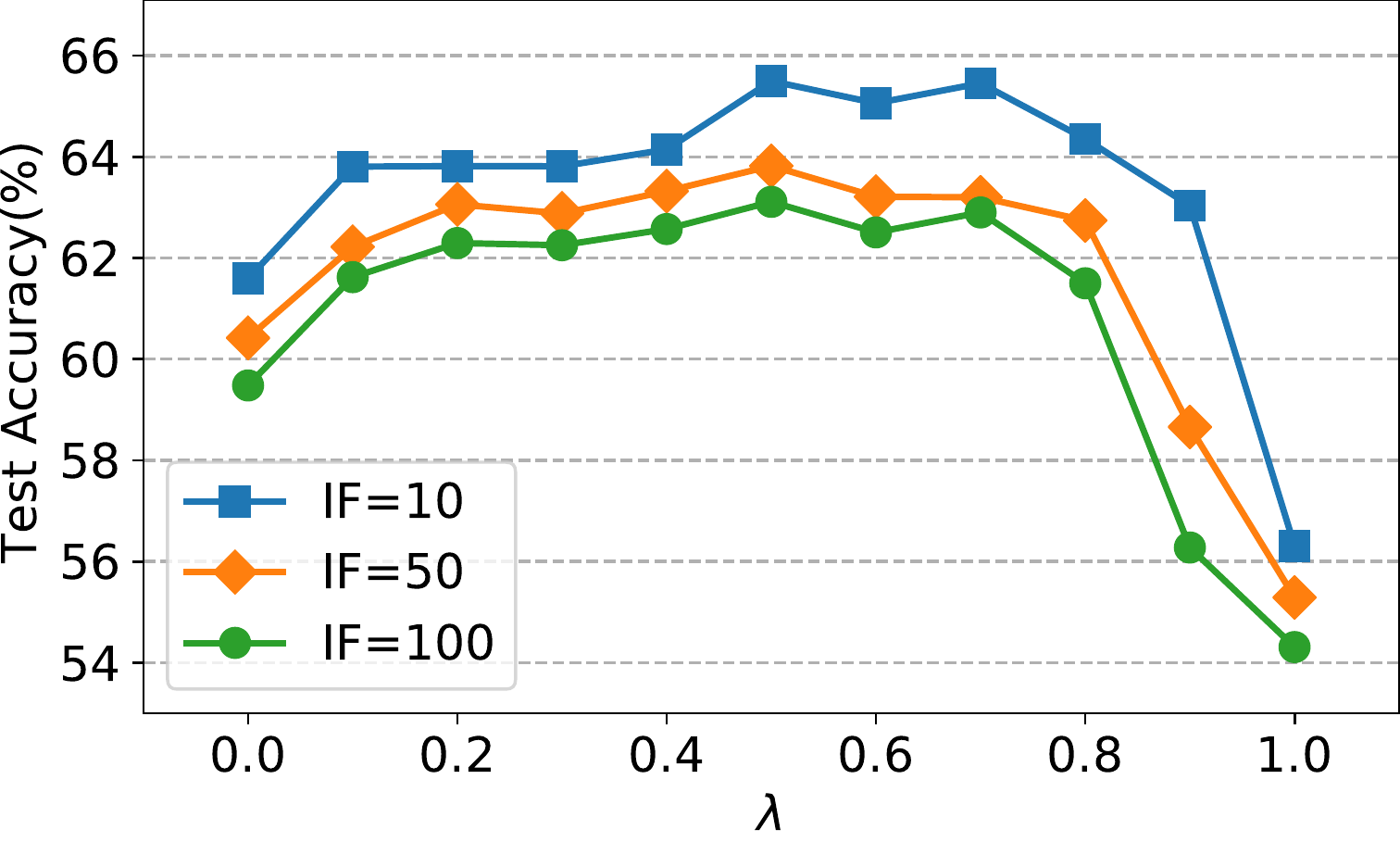}
		\caption{The performance of FEDIC with different values of $\lambda$ on CIFAR-10-LT.}
		\label{fig_lambda}
	\end{minipage}
	\hspace{0.08in} 
	\begin{minipage}[t]{0.31\linewidth}
		\centering
		\includegraphics[width=\linewidth]{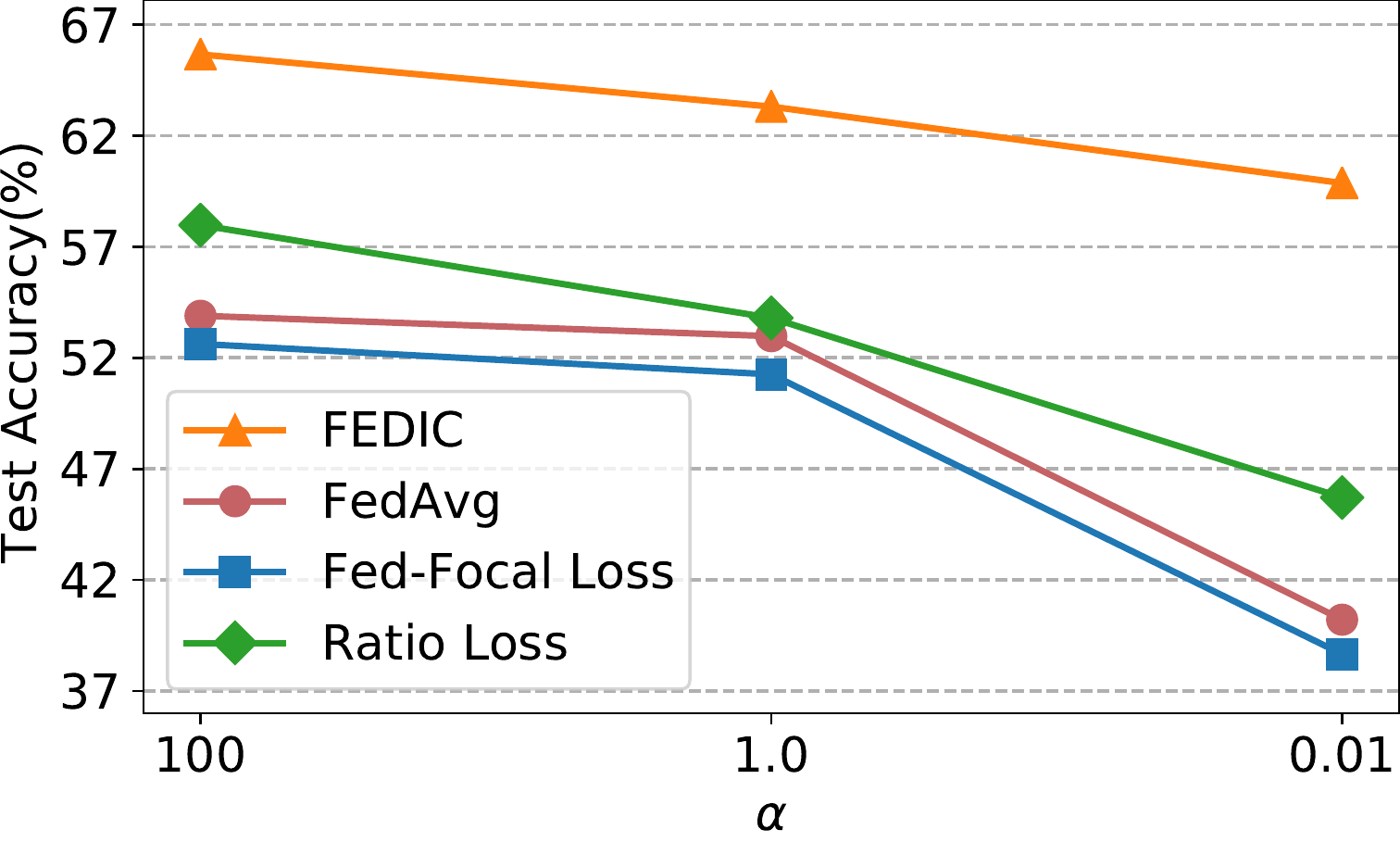}
		\caption{The performance of FEDIC with different degrees of non-IIDness on CIFAR-10-LT with IF=100.}
		\label{fig_non_iid}
	\end{minipage}
\end{figure*}
We conduct the experiments on the following datasets:

\noindent\textbf{CIFAR-10/100-LT} \cite{2009Learning}. We first exclude the auxiliary data $\DD_{aux}$ from the training data and then follow \cite{NEURIPS2019_621461af} to shape the rest of the data into a long-tail distribution with different imbalance factors (IF), which is calculated by the ratio between the number of samples in the largest class and that in the smallest class. For the unlabeled dataset $\DD_{ulb}$, we use CIFAR-100 for CIFAR-10-LT, and use the downsampled ImageNet (image size 32) for CIFAR-100-LT.

\noindent\textbf{ImageNet-LT} is a long-tailed version of ImageNet \cite{2015ImageNet}.
It contains 115.8K images from 1,000 categories, with the largest and smallest categories containing 1,280 and 5 images, respectively. We obtain the auxiliary data $\DD_{aux}$ from the balanced evaluation data and we use the oversampled CIFAR100 (image size 224) as $\DD_{ulb}$.

We use ResNet-8 for CIFAR-10-LT and CIFAR-100-LT, and ResNet-50 for ImageNet-LT as the backbone network. 
By default, we run 200 global communication rounds, with 20 clients in total and an active user ratio $C=40\%$ in each round. For local training, the batch size is set at 128 with learning rate 0.1 and SGD as the optimizer. For server training, we set the calibration steps $I$ at 100, the distillation steps $J$ at 100, and Adam with a learning rate $0.001$ is used for knowledge distillation.
Following \cite{NEURIPS2020_18df51b9}, we use Dirichlet distribution to generate the non-IID data partition among clients with the concentration parameter $\alpha=0.1$. 

\vspace{-5px}
\subsection{Comparison with the State-of-the-art Methods} \label{sota}
\vspace{-5px}
To verify the effectiveness of FEDIC, we compare the proposed method with the following federated learning (FL) methods: FedAvg \cite{mcmahan2017communication}, FedAvgM \cite{hsu2019measuring}, FedProx \cite{MLSYS2020_38af8613} and FedNova \cite{NEURIPS2020_564127c0}, and distillation-based FL methods, including FedDF \cite{NEURIPS2020_18df51b9} and FedBE \cite{chen2021fedbe}. All of them aim at producing a good global model on non-IID data. Moreover, we also compare the imbalance-oriented FL methods: Fed-Focal Loss \cite{sarkar2020fed}, Ratio Loss \cite{wang2021addressing}, and FedAvg with post-hoc methods like cRT , $\tau$-norm and LWS \cite{Kang2020Decoupling}. 

\noindent\textbf{Results on CIFAR-10/100-LT.} The results are summarized in Table \ref{t1}. FEDIC achieves the highest test accuracy on both datasets with different IFs. Compared with the baseline FedAvg, the performance gain of FEDIC is the highest when IF=100 (around 11\% for CIFAR-10-LT and 7.8\% for CIFAR-100-LT). It shows the generalization ability of FEDIC when the universal class distribution is highly long-tailed. FedAvgM, FedProx and FedNova perform similarly to FedAvg because they only deal with data non-IIDness without taking global imbalanced class distribution into account. For the distillation-based methods, FedDF and FedBE perform even worse than FedAvg. A plausible reason is that their effectiveness is based on the power of the ensemble model as the teacher model to transfer knowledge. However, the ensemble model may perform even worse than the global model on the tail classes leading to a worse distilled model due to the global imbalanced distribution. This observation also validates the necessity of ensemble calibration in FEDIC. For the imbalance-oriented FL methods, some of them (e.g., Ratio Loss) perform well in some cases compared with FedAvg. However, there is still a performance gap compared with FEDIC because they only alleviate the imbalance problem on the server but ignore the data non-IID problem.

\begin{table}[!t]\footnotesize
	\centering
	\vspace{15px}
	\caption{Top-1 test accuracy ($\%$) for FEDIC and compared FL methods on ImageNet-LT.}
	\begin{tabular}{@{}llccc@{}}
		\toprule
		& \multicolumn{4}{c}{ImageNet-LT} \\ \cmidrule(l){2-5} 
		\multirow{-2}{*}{\textbf{Method}} & All  & Many  & Medium & Few \\ \midrule
		FedAvg             &  23.85  &  34.92  &  19.18   &  7.41 \\
		FedAvgM              &  22.57  &  33.93  &  18.55   & 6.73  \\
		FedProx        &22.99      & 34.25   &  17.06  &     6.37   \\ \midrule
		FedDF               & 21.63   &  31.78  &  15.52   &  4.48 \\
		Ratio Loss            &  24.32  &   36.33 &   18.14  &  7.10 \\
		FedAvg+LWS                &  21.58  &  31.66  &  15.76   & 5.33  \\ \midrule
		\rowcolor[HTML]{EFEFEF} 
		FEDIC         &  \textbf{28.93}  &  \textbf{38.24}  &   \textbf{25.28} & \textbf{15.91}  \\ \bottomrule
	\end{tabular}
	\label{t2}
	\vspace{-10px}
\end{table}
\noindent\textbf{Results on ImageNet-LT.} We evaluate FEDIC on ImageNet-LT whose results are reported in Table \ref{t2}. 
Compared with the other methods, FEDIC achieves the best results on all cases we have tried thus far. At the same time, the accuracy on the few-shot classes achieves 15.91\%, which is a significant improvement of 8.5\% in comparison with the baseline.

\vspace{-5px}
\subsection{Model Validation}
\vspace{-5px}
\noindent\textbf{Ablation study on ensemble calibration.} We conduct an ablation study to evaluate the necessity of each component of ensemble calibration in FEDIC, as shown in Table \ref{ablation_table}. We evaluate three modules: Fine-tune (FT), client-wise and class-wise logit adjustment (LA) and knowledge distillation (KD). Note that we do not specifically evaluate the proposed calibration gating network because it is used only if both FT and LA are activated. The experiment is done by running FEDIC to round 200 and evaluating all combinations on that round.
In the upper part (a)-(d) in Table \ref{ablation_table}, we only evaluate the performance of the ensemble model without distillation. Compared with the baseline (a), the overall accuracy of the calibrated ensemble model (d) is improved by 11.2\%. 
In the lower part (e)-(h), knowledge distillation is conducted. It can be observed that 
the gap of the overall accuracy between the teacher model (d) and the student model (h) is only 1.7\%, which indicates that the generalization ability to deal with the long-tail distribution is successfully transferred.
\begin{table}[!t]\footnotesize
	\vspace{15px}
	\centering
	\caption{Ablation study on the components of ensemble calibration in FEDIC on CIFAR-10-LT with IF=100.}
	\begin{tabular}{@{}cccccccc@{}}
		\toprule
		\multicolumn{4}{c}{\textbf{Module}} & \multicolumn{4}{c}{CIFAR-10-LT} \\ \midrule
		&FT   & LA   & KD   & All & Many & Medium & Few \\ \midrule
		(a)&\xmark    & \xmark    & \xmark   &   53.59  & \textbf{76.90}   &  50.12  & 34.90  \\
		(b)&\cmark    & \xmark    & \xmark   &  61.11   &  64.30  &  62.47  &  56.10 \\
		(c)&\xmark    & \cmark    & \xmark   & 60.66    & 66.00   &  60.93   & 53.26  \\
		(d)&\cmark    & \cmark    & \xmark   &  \textbf{64.77}   &  65.26  &  \textbf{64.88}   &  \textbf{62.55} \\ \midrule
		(e)&\xmark    & \xmark    & \cmark   &  53.92    &  \textbf{77.20}  &  51.32   & 34.05  \\		
		(f)&\cmark    & \xmark    & \cmark   &  60.18  & 64.68  &  58.04  & 56.32  \\
		(g)&\xmark    & \cmark    & \cmark   &  61.71   &  63.75  &  63.45   & 57.94  \\
		(h)&\cmark    & \cmark   & \cmark   &   \textbf{63.11}   &  64.90  &  \textbf{63.60}  & \textbf{60.83}  \\ \bottomrule
	\end{tabular}
	\label{ablation_table}
	\vspace{-10px}
\end{table}

\noindent\textbf{Influence of sizes of auxiliary and unlabeled datasets.} The sizes of $\DD_{aux}$ and $\DD_{ulb}$ play a key role in FEDIC. Therefore, we evaluate its influence on the performance of FEDIC, compared with the baseline FedAvg and a global model fine-tuned with $\DD_{aux}$ (marked as FT), as shown in Fig. \ref{fig_unlabeled}. Different curves in the figure indicate the data fractions of the unlabeled data used for distillation. We observe that FEDIC consistently outperforms FedAvg and FT for all sizes of $\DD_{aux}$. 

\noindent\textbf{Sensitivity analysis of hyperparameters.} We investigate the impact of distillation trade-off coefficient $\lambda$. This hyperparameter controls the strength of distillation in the loss function in Eq. (\ref{distill}). It can be observed from Fig. \ref{fig_lambda} that FEDIC is robust to most $\lambda$ values. However, the performance severely drops when $\lambda$ reaches 1, which shows that solely distillation with unlabeled data is not enough for a good global model. 

\noindent\textbf{Influence of the degree of non-IIDness.} Fig. \ref{fig_non_iid} further shows the test accuracy of four methods under the different degrees of non-IIDness. It can be observed that the performance of all methods drops as the degree of non-IIDness increases. However, the performance of the compared methods drops more severely than FEDIC when $\alpha$ decreases from 1.0 to 0.01. 
\vspace{-5px}
\section{Conclusion}
\vspace{-5px}
In this paper, we have proposed FEDIC to deal with the problem of learning a global model on non-IID and long-tailed data in the federated learning framework. FEDIC is a server-side method that first calibrates the biased ensemble model against the long-tail distribution by client-wise and class-wise logit adjustment with a calibration gating network. Then, the calibrated ensemble is used as the teacher model to transfer knowledge to the global model for further optimization on the clients. Also, the effectiveness of each component in FEDIC has been validated empirically. Experiments have shown that FEDIC outperforms the state-of-the-art FL methods on datasets with the non-IID and long-tailed setting. 
\vspace{-5px}
\bibliographystyle{IEEEbib}\small
\bibliography{icme2022template}

\end{document}